\documentclass[runningheads]{llncs}
\usepackage{graphicx}
\usepackage{adjustbox}
\usepackage{booktabs}
\usepackage{subcaption}
\usepackage{amsmath}
\begin{document}
\title{ICDAR 2021 Competition on \\ On-Line Signature Verification}
%
%
\author{Ruben Tolosana\inst{1} \and Ruben Vera-Rodriguez\inst{1} \and Carlos Gonzalez-Garcia\inst{1} \and Julian Fierrez\inst{1} \and Santiago Rengifo\inst{1} \and Aythami Morales\inst{1} \and Javier Ortega-Garcia\inst{1} \and Juan Carlos Ruiz-Garcia\inst{1} \and Sergio Romero-Tapiador\inst{1} \and Jiajia Jiang\inst{2} \and Songxuan Lai\inst{2} \and Lianwen Jin\inst{2,3} \and Yecheng Zhu\inst{2} \and Javier Galbally\inst{4} \and
Moises Diaz\inst{5} \and Miguel Angel Ferrer\inst{6} \and Marta Gomez-Barrero\inst{7} \and Ilya Hodashinsky\inst{8} \and Konstantin Sarin\inst{8} \and Artem Slezkin\inst{8} \and Marina Bardamova\inst{8} \and Mikhail Svetlakov\inst{8} \and Mohammad Saleem\inst{9} \and Cintia Lia Szücs\inst{9} \and Bence Kovari\inst{9} \and Falk Pulsmeyer\inst{10} \and Mohamad Wehbi\inst{10} \and Dario Zanca\inst{10} \and Sumaiya Ahmad\inst{11} \and Sarthak Mishra\inst{11} \and Suraiya Jabin\inst{11}}
\authorrunning{R. Tolosana et al.}
%
\institute{Biometrics and Data Pattern Analytics Lab, UAM, Spain \\
\email{(ruben.tolosana;ruben.vera;julian.fierrez;aythami.morales;javier.ortega)@uam.es\\
(santiago.rengifo;juanc.ruiz)@uam.es (carlos.gonzalezgarcia;sergio.romerot)@estudiante.uam.es}
\and
South China University of Technology, China\\
\and
Guangdong Artificial Intelligence and Digital Economy Laboratory, China\\
\email{eejiajia\_jiang@mail.scut.edu.cn;eesxlai@qq.com;\\eelwjin@scut.edu.cn;claytonzyc@foxmail.com}
\and
European Commission - Joint Research Centre, Italy\\
\email{javier.galbally@ec.europa.eu}
\and
Universidad del Atlantico Medio, Spain\\
\email{moises.diaz@atlanticomedio.es}
\and
Universidad de las Palmas de Gran Canaria, Spain\\
\email{miguelangel.ferrer@ulpgc.es}
\and
Hochschule Ansbach, Germany\\
\email{marta.gomez-barrero@hs-ansbach.de}
\and
Tomsk State University of Control Systems and Radioelectronics, Russia\\
\email{hodashn@rambler.ru;sks@security.tomsk.ru;\\saotom724@gmail.com;722bmb@gmail.com;rvvincle@gmail.com}
\and
Budapest University of Technology and Economics, Hungary\\
\email{(Mohammad.Saleem;Szucs.CintiaLia;kovari)@aut.bme.hu}
\and
Machine Learning and Data Analytics Lab, FAU, Germany\\
\email{(falk.pulsmeyer;mohamad.wehbi;dario.zanca)@fau.de}
\and
Jamia Millia Islamia, India
\email{sumaiya.hld@gmail.com;sarthak.mishra2905@gmail.com;sjabin@jmi.ac.in}
}

\maketitle              
\begin{abstract}
This paper describes the experimental framework and results of the ICDAR 2021 Competition on On-Line Signature Verification (SVC 2021). The goal of SVC 2021 is to evaluate the limits of on-line signature verification systems on popular scenarios (office/mobile) and writing inputs (stylus/finger) through large-scale public databases. Three different tasks are considered in the competition, simulating realistic scenarios as both random and skilled forgeries are simultaneously considered on each task. The results obtained in SVC 2021 prove the high potential of deep learning methods. In particular, the best on-line signature verification system of SVC 2021 obtained Equal Error Rate (EER) values of 3.33\% (Task 1), 7.41\% (Task 2), and 6.04\% (Task 3).

SVC 2021 will be established as an on-going competition\footnote{\url{https://sites.google.com/view/SVC2021}}, where researchers can easily benchmark their systems against the state of the art in an open common platform using large-scale public databases such as DeepSignDB\footnote{\url{https://github.com/BiDAlab/DeepSignDB}} and SVC2021\_EvalDB\footnote{\url{https://github.com/BiDAlab/SVC2021\_EvalDB}}, and standard experimental protocols.

\end{abstract}
\section{Introduction}
On-line handwritten signature verification has always been a very active area of research due to its high popularity for authentication scenarios~\cite{moises_ACM} and the variety of open challenges that are still under research nowadays~\cite{2020_Signature_CognitiveComputation}, e.g., one/few-shot learning~\cite{galbally09ICDARenrollment,moises_1signature,lai2020synsig2vec,2021_AAAI_DeepWriteSYN}, device interoperability~\cite{Alonso-Fernandez2005_TabletPC,Marcos08a,sae2014online,2015_IEEEAccess_InterSign_Tolosana}, aging~\cite{galbally13PONEagingSignature,2019_IETB_Aging_Tolosana}, types of impostors~\cite{2018_HanbookBioAntiSpoofing_signature_Tolosana,galbally2017accuracy}, signature complexity~\cite{Sonia_client_entropy_EER,tolosana2019exploiting,Vera_2019_Complexity}, template storage~\cite{delgado2019biometric}, etc. Despite all these challenges, the performance of on-line signature verification systems has been improved in the last years due to several factors, especially: \textit{i)} the evolution in the acquisition technology going from devices specifically designed to acquire handwriting and signature in office-like scenarios through a pen stylus (e.g. Wacom devices) to the current touch screens of mobile scenarios in which signatures can be captured anywhere using our own personal smartphone through the finger~\cite{sae2014online,eBioSign_journal,antal2018online,ellavarason2020touch}, and \textit{ii)} the extended usage of deep learning technology in many different areas, overcoming traditional handcrafted approaches and even human performance~\cite{2021_TBIOM_DeepSign_Tolosana,2018_IEEEAccess_RNN_Tolosana,Lai_TIFS_2018,Uchida2019_DDTW,ahrabian2018usage}. So, with all these aspects in mind, the question is: what are the current performance limits of the on-line signature verification technology under realistic scenarios? 

This paper describes the experimental framework and results of the ICDAR 2021 Competition on On-Line Signature Verification (SVC 2021). The goal of SVC 2021 is to evaluate the limits of on-line signature verification systems on popular scenarios (office/mobile) and writing inputs (stylus/finger) through large-scale public databases. Three different tasks are considered in the competition:

\begin{itemize}
\item \textbf{Task 1}: Analysis of office scenarios using the stylus as input.
\item \textbf{Task 2}: Analysis of mobile scenarios using the finger as input.
\item \textbf{Task 3}: Analysis of both office and mobile scenarios simultaneously. 
\end{itemize}

In addition, we simulate in SVC 2021 the following realistic operational conditions, to the best of our knowledge, not considered in previous on-line signature verification competitions~\cite{yeung2004svc2004,conf/icdar/BlankersHFV09,Houmani2012993,malik2013icdar,malik2015icdar2015}:

\begin{itemize}
\item Over 1,700 subjects and 100 different acquisition devices are considered in the competition, using both Wacom devices (office scenarios) and general purpose devices such as tablets and smartphones (mobile scenarios).   
\item Random and skilled forgeries are simultaneously considered in each task. In addition, different types of skilled forgeries are considered in the competition such as static (i.e., only the image of the signature to forge is available) and dynamic forgeries (i.e., both image and dynamics are available), in both trained and blueprint cases~\cite{2018_HanbookBioAntiSpoofing_signature_Tolosana}.
\item High intra-subject variability (a.k.a. aging) as different acquisition set-ups are considered in the competition ranging from 1 to 5 sessions, and with a time gap between sessions from days to months.
\end{itemize}

This realistic scenario has been achieved thanks to the public DeepSignDB database~\cite{2021_TBIOM_DeepSign_Tolosana} and the novel SVC2021\_EvalDB database (this later one specifically acquired for SVC 2021). Besides, we have designed realistic and challenging experimental protocols making public the corresponding signature comparisons files and the benchmarking platform.

SVC 2021 will be established as an on-going competition\footnote{\url{https://sites.google.com/view/SVC2021}}, where researchers can easily benchmark their systems against the state of the art using a common experimental protocol and an open computing platform (CodaLab\footnote{\url{https://competitions.codalab.org/competitions/27295}}).

The remainder of the paper is organised as follows. Sec.~\ref{Databases} and~\ref{Experimental_Protocol} describe the details of the databases and the set-up considered in the competition, respectively. Sec.~\ref{System_Description} provides a description of the submitted on-line signature verification systems. Sec.~\ref{Experimental_Results} describes the results of the competition. Finally, Sec.~\ref{Conclusions} draws the final conclusions.

\section{SVC 2021: Databases}\label{Databases}
Two databases are considered in SVC 2021: DeepSignDB and SVC2021\_EvalDB. These databases are publicly available for the research community and can be downloaded following the instructions included in\footnote{\url{https://github.com/BiDAlab/DeepSignDB}}~\footnote{\url{https://github.com/BiDAlab/SVC2021\_EvalDB}}. We provide next a description of them.

\subsection{DeepSignDB}
The DeepSignDB database~\cite{2021_TBIOM_DeepSign_Tolosana} comprises on-line signatures from a total of 1,526 subjects from four different well-known databases: MCYT (330 subjects)~\cite{Ortega_Garcia2003_MCYT},
BiosecurID (400 subjects)~\cite{Fierrez2009_PAA}, Biosecure DS2 (650 subjects)~\cite{Houmani2012993}, eBioSign (65 subjects)~\cite{eBioSign_journal}, and a novel signature database composed of 81 subjects. DeepSignDB comprises more than 70K signatures acquired using both stylus and finger writing inputs in both office and mobile scenarios. A total of 8 different devices are considered in the acquisition (i.e., 5 Wacom devices and 3 Samsung general purpose devices). In addition, different types of impostors and number of acquisition sessions are considered along the database. For more details about DeepSignDB, we refer the reader to the published article~\cite{2021_TBIOM_DeepSign_Tolosana}.

\subsection{SVC2021\_EvalDB}
The SVC2021\_EvalDB database is a novel database specifically acquired for SVC 2021. Two acquisition scenarios are considered: office and mobile scenarios.

\begin{itemize}
\item \textbf{Office scenario:} on-line signatures from 75 total subjects were acquired using a Wacom STU-530 device with the stylus as writing input. Regarding the acquisition protocol, the
device was placed on a desktop and subjects were able to rotate it in order to feel comfortable with the writing position. It is important to highlight that the subjects considered in the acquisition of SVC2021\_EvalDB are different compared to the ones considered in the DeepSignDB database.

Signatures were collected in two separated sessions with a time gap between them of at least 1 week. For each subject, there are 8 total genuine signatures (4 signatures/session) and 16 skilled forgeries (8 signatures/type) performed by four different subjects in two different sessions. Regarding the skilled forgeries, both static and dynamic forgeries were considered in the first and second acquisition sessions, respectively. Information related to \textit{X} and \textit{Y} spatial coordinates, pressure, and timestamp is recorded for the Wacom device. In addition, pen-up trajectories are also available. 

\item \textbf{Mobile scenario:} on-line signatures from 119 total subjects were acquired using the same acquisition framework considered in MobileTouchDB~\cite{2020_TIFS_BioTouchPass2_Tolosana}. Regarding the acquisition protocol, we implemented an Android App and uploaded it to the Play Store in order to study an unsupervised mobile scenario. This way all subjects could download the App and use it on their own devices without any kind of supervision, simulating a practical scenario in which subjects can generate touchscreen on-line signatures in any possible scenario, e.g., standing, sitting, walking, indoors, outdoors, etc. As a result, 94 different smartphone models from 16 different brands were collected during the acquisition.

Regarding the acquisition protocol, between four and six separated sessions in different days were considered with a total time gap between the first and last session of at least 3 weeks. For each subject, there are at least 8 total genuine signatures (2 signatures/session) and 16 skilled forgeries (8 signatures/type) performed by four different subjects. Regarding the skilled forgeries, both static and dynamic forgeries were considered, similar to the office scenario. Information related to \textit{X} and \textit{Y} spatial coordinates, and timestamp is recorded for all devices. Pen-up information is not available in this case.

\end{itemize}

\section{SVC 2021: Competition Set-Up}\label{Experimental_Protocol}

\subsection{Tasks}
The goal of SVC 2021 is to evaluate the limits of on-line signature verification systems on popular scenarios (office/mobile) and writing inputs (stylus/finger) through large-scale public databases. As a result, the following three tasks are considered in the competition: 

\begin{itemize}
\item \textbf{Task 1}: Analysis of office scenarios using the stylus as input.
\item \textbf{Task 2}: Analysis of mobile scenarios using the finger as input.
\item \textbf{Task 3}: Analysis of both office and mobile scenarios simultaneously. 
\end{itemize}

In addition, SVC 2021 simulates realistic operational conditions considering random and skilled forgeries simultaneously in each task. 

\subsection{Evaluation Criteria}\label{evaluation_criteria}
The SVC 2021 competition follows a ranking based on points. Each task is evaluated separately, having three winners with their corresponding points (gold medal: 3, silver medal: 2, and bronze medal: 1). The participant/team that gets more points in total (Task 1, 2, and 3) in the final evaluation stage of the competition is the winner of SVC 2021.

The evaluation metric considered is the popular Equal Error Rate (\%) similar to most on-line signature verification studies in the literature.

\subsection{Experimental Protocol}\label{experimental_protocol}
The two following stages are considered in SVC 2021: 

\begin{itemize}
\item \textbf{Development:} the goal of this stage is to provide the participants with the data needed to train the on-line signature verification systems. Only the DeepSignDB database is provided to the participants in this stage of the competition. In addition, participants can freely use other databases to train their systems.

In order to allow the participants to test their trained systems under similar conditions considered in the final evaluation stage of the competition, we divide the DeepSignDB database into training and evaluation datasets. The training dataset is based on 1,084 subjects whereas the evaluation dataset comprises the remaining 442 subjects of the database. For the training of the systems (1,084 subjects), no instructions are given to the participants. They can use the data as they like. Nevertheless, for the evaluation of the systems (442 subjects), we provide the participants with the signature comparisons to run. Participants can run their on-line signature verification systems using the signature comparisons files provided to obtain the scores and test the EER performance on the public web platform (CodaLab) created for the competition\footnote{\url{https://competitions.codalab.org/competitions/27295}}. This way participants can obtain a quantitative measure of the performance of the developed systems for the final evaluation stage of the competition. 

In this development stage of the competition, participants can submit up to 300 system evaluation trials in total for all three tasks together. Results are updated in CodaLab in real time and they are visible to everyone in a ranking dashboard.  

\item \textbf{Final Evaluation:} the final evaluation of SVC 2021 is carried out using only the novel SVC2021\_EvalDB database. The database together with the corresponding signature comparisons files (one file per task) are sent to the participants after signing the corresponding license agreement. It is important to highlight that all signatures are included in a single folder, and both the nomenclature of the signatures and the signature comparisons files are randomized to avoid cheating. Ground-truth labels are not provided to the participants. In addition, and in order to consider a very challenging impostor scenario, the skilled forgery comparisons included in the corresponding files are optimised using machine learning methods, selecting only the best high-quality forgeries.

In this final evaluation of SVC 2021, participants are allowed to submit the scores achieved by up to 3 different signature verification systems for each of the tasks considered in the competition.
    
\end{itemize}

\section{SVC 2021: Description of Evaluated Systems}\label{System_Description}
A total of 54 participants/teams initially registered in SVC 2021. However, only 6 teams finally submitted their scores with a total of 12 different on-line signature verification systems. Next, we describe briefly the systems provided by each of the teams of the competition.

\subsection{DLVC-Lab Team}
The DLVC-Lab team is composed of members of the South China University of Technology, and the Guangdong Artificial Intelligence and Digital Economy Laboratory.

The DLVC-Lab team proposed an end-to-end trainable deep soft-DTW (DSDTW) model, which greatly enhances the classical Dynamic Time Warping (DTW) method with the capability of deep representation learning. In particular, they use neural networks to learn deep time functions as inputs for DTW. As DTW is not fully differentiable with regards to its inputs, they introduce its smoothed formulation, soft-DTW~\cite{cuturi2017soft}, and incorporate the soft-DTW distances of signature pairs into a triplet loss function for optimization. As soft-DTW is differentiable, the entire system is end-to-end trainable and achieves a perfect integration of neural networks and DTW. 

Three different approaches were submitted to SVC 2021. System 1 is based on Convolutional Recurrent Neural Networks (CRNN) whereas System 2 and 3 are based on fully Convolutional Neural Networks (CNN). Systems 2 and 3 only differ in the training data. Concretely, Systems 1 and 2 use the development set of the DeepSignDB database for training (1,084 subjects), including both stylus-written and finger-written signatures. System 3 uses only finger-written signatures for training. 

Regarding the feature extraction, 12 total time functions are extracted for each signature, considering information such as velocity and acceleration. These time functions are fed to the DSDTW.

\subsection{SIG Team}
The Spanish-Italian-German (SIG) team is composed of members of the European Commission (Italy), Universidad del Atlantico Medio (Spain), Universidad de las Palmas de Gran Canaria (Spain), and Hochschule Ansbach (Germany).

The signature verification system presented is based on the main principle laid out in~\cite{galbally2015line}: the generation of synthetic off-line signatures from the real on-line samples and the fusion of both types of data can lead to the overall improvement of the on-line verification performance. Following that rationale, the system submitted is based on the combination of on- and off-line signature information. 

The \textbf{on-line signature approach} is based on local features and the well-known DTW algorithm. In particular, the system is based on a subset of the initial 27 time functions introduced in \cite{martinez14mobileSignature} and selected using the Sequential Floating Forward Selection (SFFS) algorithm. The specific implementation of the DTW algorithm uses the Euclidean Distance to compute the optimal path in between signatures and outputs as score $s_{\mathrm{on}}$ the last value of the optimal path, normalised by the path length. Please be aware that, for the cases where pressure $p$ is not available (i.e., mobile scenario in Task 2 of the competition), the time signal is simply discarded, together with any other time function derived from it.

Regarding the \textbf{off-line signature approach}, the first step performed is the generation of the synthetic off-line data starting from the real on-line signatures. Two different methods are used for this purpose: \textit{i)} continuous trace~\cite{diaz2014generation,galbally2015line}, and \textit{ii)} dotted trace~\cite{diaz2019dynamically}. Once the two synthetic off-line signatures are created (for each dynamic signature given as input), three different handcrafted features are extracted: \textit{i)} run-length distribution~\cite{bouamra2018towards}, \textit{ii)} geometrical features~\cite{ferrer2005offline}, and \textit{iii)} quad-tree implementation of histogram of templates~\cite{serdouk2017handwritten}. 

The score for each of the three feature sets is obtained by comparing the reference and probe vectors using the DTW algorithm followed by the cityblock distance. This process leads to six off-line intermediate scores ($s1_{\mathrm{off}}$, $s2_{\mathrm{off}}$,..., $s6_{\mathrm{off}}$) for each on-line comparison defined in the competition (recall that each individual on-line signature is converted to two off-line synthetic signatures, defined by three different feature sets).

The six intermediate scores obtained by the off-line approach are finally fused into one unique off-line score $s_{\mathrm{off}}$ using a weighted sum. The weights for the fusion are empirically calculated on the training databases of the competition optimising the EER for each of the tasks considered in the assessment.

Finally, the on- and off-line scores ($s_{\mathrm{on}}$ and $s_{\mathrm{off}}$) are normalised to the [0,1] range using the tanh-estimators and fused into the final score $s$ given as output by the system based on the weighted sum.

Only the DeepSignDB database provided in the development stage of SVC 2021 was considered for training and evaluating the system.

%
%
%
%
%
%
%
%

\subsection{TUSUR KIBEVS Team}
The TUSUR KIBEVS team is composed of members of the Tomsk State University of Control Systems and Radioelectronics.

The on-line signature verification system presented is based on the use of global features and a gradient boosting classifier. First, a set of 100 global features is extracted for each enrolled and test signatures ($F_{enrolled}$ and $F_{test}$) based on previous approaches in the literature~\cite{Fierrez-Aguilar2005_AVBPA_changed}. Then, a new feature vector $F$ is obtained based on the subtraction of the previous enrolled and test feature vectors: $F=| F_{enrolled} - F_{test}|$. The resulting feature vector $F$ is introduced to CatBoost~\cite{prokhorenkova2018catboost}, a fast, scalable, and high performance Gradient Boosting on Decision Trees (GBDT) that is available as an open source library\footnote{\url{https://catboost.ai/}}.

Regarding the training procedure, only the DeepSignDB database provided in the development stage of SVC 2021 is considered. A total of 10K signature comparisons are randomly selected (5K genuine and 5K forgeries), considering both office (stylus) and mobile (finger) scenarios simultaneously. Forgery comparisons included 2.5K skilled forgeries and 2.5K random forgeries. 

\subsection{SigStat Team}
The SigStat team is composed of members of the Budapest University of Technology and Economics.

Three different on-line signature verification systems were presented. All of them are implemented using the SigStat framework\footnote{\url{http://www.sigstat.org}}. First, all signatures go through a preprocessing stage. Time samples with zero pressure are removed from the stylus-based signatures to reduce noise and remove some artifacts. Finally, \textit{X}, \textit{Y}, and pressure information are scaled to the [0,1] range and shifted by the average of their values. After this preprocessing stage, the biometric information is used to calculate different distance scores between signature pairs, considering three different approaches.

The first system considers local thresholds to detect whether the query signature is genuine of forgery. In particular, it uses DTW to calculate signature distances and the k-Nearest Neighbours (k-NN) approach to set a lower and an upper threshold for each reference signature. During the development stage, the system is tested on the evaluation subset of the DeepSignDB (442 users). The distances and comparisons between the signatures are used to calculate and tune several parameters, selecting the optimal values of the genuine $G_{th}$ and forgery $F_{th}$ thresholds and a scaling parameter $s$ for the classification purpose. 

For testing, the distance $d$ between the questioned signature $(S_q)$ and the reference signature $(S_r)$ is obtained using DTW. The final score $P_q$ is calculated as follows:

\begin{equation}
    P_q = \dfrac{s \cdot F_{th} - d}{s\cdot F_{th} - G_{th}}
\end{equation}

The second system considers global thresholds and is based on 4 classifiers and a linear fusion of them. The first three classifiers take advantage of global features such as the standard deviation of \textit{X} and \textit{Y} spatial coordinates, and the signing time duration. The last classifier is based on the DTW distance of signature pairs. 

In the development stage, the evaluation subset of DeepSignDB is used to make genuine-genuine and genuine-forgery comparisons. For each comparison, the calculated DTW distance, the device input, and the expected prediction are stored. Next, the comparisons and their results are sorted into four different groups based on expected prediction and input device (genuine finger, genuine stylus, forgery finger, and forgery stylus). For each group some statistical parameters such as the minimum and median values are calculated and used to set the global thresholds for the system.

For testing, the score of the questioned signature $P_q$ is calculated based on the DTW distance of the reference-questioned pair $d$, the minimum distance of genuine comparisons $d_{g_{min}}$ and the median distance of forgery comparisons $d_{f_{med}}$:

\begin{equation}
    P_q = 1 - \dfrac{d_{f_{med}} - d}{d_{f_{med}} - d_{g_{min}}}
\end{equation}

In case of $d < d_{g_{min}}$, the score $P_q$ is automatically 0 and when $d > d_{f_{med}}$ is 1. A similar approach is considered for the remaining three classifiers based on global features. 

Finally, the third system extends the set of global features considered in the second system, for example including the DTW distance as feature. Contrary to previous systems, a gradient boosting classifier (XGBoost) is considered for the final prediction.

\subsection{MaD-Lab Team}
The MaD-Lab team is composed of members of the Machine Learning and Data Analytics Lab (FAU).

The proposed system consists of a 1D CNN trained to classify pairs of signatures as matching or not matching. Features are extracted using a mathematical concept called \textit{path signature} together with statistical features. These features are then used to train an adapted version of ResNet-18~\cite{he2016deep}.

Regarding the preprocessing stage, the \textit{X} and \textit{Y} spatial coordinates are normalised to a $\left[ -1, 1\right]$ range whereas the pressure information to $[0,1]$. In case that no pressure information is available (Task 2, mobile scenario), a vector with all one values is considered.

For the feature extraction, a set of global features related to statistical information is extracted for each signature. Besides, additional features are extracted using the signature path method~\cite{chevyrev2016primer}. This is a mathematical tool that extracts features from paths. It is able to encode linear and non-linear features from the signature path. The path signature method is applied over the raw \textit{X} and \textit{Y} spatial coordinates, their first-order derivatives, the perpendicular vector to the segment, and the pressure. 

Finally, for classification, a 1D adapted version of the ResNet-18 CNN is considered. To adapt the ResNet-18 image version, every 2D operation is exchanged with a 1D one. Also, a sigmoid activation function is added in the last layer to output values between $0$ and $1$. Pairs of signatures are presented to the network as two different channels. 

Regarding the training parameters of the network, binary cross-entropy is used as the loss function. The network is optimised using stochastic gradient descent (SGD) with a momentum of $0.9$ and a learning rate of $0.001$. The learning rate is decreased by a factor $0.1$ if the accumulated loss in the last epoch is larger than the epoch before. In case the learning rate drops to $10^{-6}$, the training process is stopped. Also, if the learning rate does not decrease below $10^{-6}$, the training process is stopped after $50$ epochs.

\subsection{JAIRG Team}
The JAIRG team is composed of members of the Jamia Millia Islamia.

Three different systems were presented, all of them focused on Task 2 (mobile scenarios). The on-line signature verification systems considered are based on an ensemble of different deep learning models training with different sets of features. The ensemble is formed using a weighted average of the scores provided by five individual systems. The specific weights to fuse the scores in the ensemble approach are obtained using a Genetic Algorithm (GA)~\cite{Galbally2007_GeneticFS}.

For the feature extraction, three different approaches are considered: \textit{i)} a set of 18 time functions related to \textit{X} and \textit{Y} spatial coordinates~\cite{2015_IEEEAccess_InterSign_Tolosana}, \textit{ii)} a subset of 40 global features~\cite{martinez14mobileSignRobustPerf}, and \textit{iii)} a set of global features extracted after applying 2D Discrete Wavelet Transform (2D-DWT) over the image of the signatures.

For classification, Bidirectional Gated Recurrent Unit (BGRU) models with a Siamese architecture are considered~\cite{2018_IEEEAccess_RNN_Tolosana}. Different models are studied varying the number of hidden layers, input features, and training parameters. Finally, an ensemble of the best BGRU models in the evaluation of DeepSignDB is considered, selecting the fusing weight parameters through a GA.

\begin{table}[t]
\centering
\caption{\textbf{Final evaluation results of SVC 2021} using the novel SVC2021\_EvalDB database acquired for the competition. For each specific task, we include the points achieved by each team depending on the ranking position (gold medal: 3, silver medal: 2, and bronze medal: 1).}\vspace{4mm}
\scalebox{0.8}{
\begin{tabular}{ccc|ccc|ccc}
\multicolumn{3}{c|}{\textbf{Task 1: Office Scenario}} & \multicolumn{3}{c|}{\textbf{Task 2: Mobile Scenario}} & \multicolumn{3}{c}{\textbf{Task 3: Office/Mobile Scenario}} \\ \hline
\textit{Points}  & \textit{Team}  & \textit{EER(\%)}  & \textit{Points}  & \textit{Team}  & \textit{EER(\%)}  & \textit{Points}     & \textit{Team}    & \textit{EER(\%)}    \\ \hline
3                & DLVC-Lab       & 3.33\%            & 3                & DLVC-Lab       & 7.41\%            & 3                   & DLVC-Lab         & 6.04\%              \\
2                & TUSUR KIBEVS   & 6.44\%            & 2                & SIG            & 10.14\%           & 2                   & SIG              & 9.96\%              \\
1                & SIG            & 7.50\%            & 1                & SigStat        & 13.29\%           & 1                   & TUSUR KIBEVS     & 11.42\%             \\ \hline
0                & MaD            & 9.83\%            & 0                & TUSUR KIBEVS   & 13.39\%           & 0                   & MaD              & 14.21\%             \\
0                & SigStat        & 11.75\%           & 0                & Baseline DTW   & 14.92\%           & 0                   & SigStat          & 14.48\%             \\
0                & Baseline DTW   & 13.08\%           & 0                & MaD            & 17.23\%           & 0                   & Baseline DTW     & 14.67\%             \\
                  &                &                   & 0                & JAIRG          & 18.43             &                     &                  &                    
\end{tabular}
}
\label{Table_Evaluation}
\end{table} 

\begin{table}[t]
\centering
\caption{\textbf{Global ranking of SVC 2021.}}\vspace{3mm}
\scalebox{0.85}{
\begin{tabular}{c|c|c}
\textbf{Position} & \textbf{Team} & \textbf{Total Points} \\ \hline
\textbf{1}                 & \textbf{DLVC-Lab}      & \textbf{9}                     \\
2                 & SIG           & 5                     \\
3                 & TUSUR KIBEVS  & 3                     \\
4                 & SigStat       & 1                     \\
5                 & MaD           & 0                     \\
6                 & JAIRG         & 0                    
\end{tabular}
}
\label{Table_Ranking}
\end{table}

\section{SVC 2021: Experimental Results}\label{Experimental_Results}
This section describes the final evaluation results of the competition using the novel SVC2021\_EvalDB database acquired for SVC 2021. It is important to highlight that the winner of SVC 2021 is based only on the results achieved in this stage of the competition as described in Sec.~\ref{experimental_protocol}. Tables~\ref{Table_Evaluation} and~\ref{Table_Ranking} show the results achieved by the participants in each of the three tasks, and the final ranking of SVC 2021 based on the total points, respectively. For completeness, we include in Table~\ref{Table_Evaluation} a Baseline DTW system (similar to the one described in~\cite{Marcos_matching}) based on \textit{X}, \textit{Y} spatial coordinates, and their first- and second-order derivatives for a better comparison of the results.

As can be seen in Tables~\ref{Table_Evaluation} and~\ref{Table_Ranking}, DLVC-Lab is the winner of SVC 2021 (9 points), followed by SIG (5 points) and TUSUR KIBEVS (3 points). It is important to highlight that the on-line signature verification systems proposed by DLVC-Lab achieve the best results in all three tasks. In particular, an EER absolute improvement of 3.11\%, 2.73\%, and 3.92\% is achieved in each of the tasks compared to the results obtained by the second-position team. Also, it is interesting to compare the best results achieved in each task with the results obtained using traditional approaches in the field (Baseline DTW). Concretely, for each of the tasks, DLVC-Lab achieves relative improvements of 74.54\%, 50.34\%, and 58.3\% EER compared to the Baseline DTW. These results prove the high potential of deep learning approaches for the on-line signature verification field, as commented in previous studies~\cite{2021_TBIOM_DeepSign_Tolosana,2021_AAAI_DeepWriteSYN}.

Other approaches like the ones presented by the SIG team based on the use of on- and off-line signature information have provided very good results, achieving points in all three tasks (5 total points). The same happens with the system proposed by the TUSUR KIBEVS team based on global features and a gradient boosting classifier (CatBoost~\cite{prokhorenkova2018catboost}), achieving 3 points in total. In particular, the approach presented by TUSUR KIBEVS has outperformed the approach proposed by the SIG team for the office scenario (6.44\% vs. 7.50\% EER). Nevertheless, much better results are obtained by the SIG team for the mobile and office/mobile scenarios (10.14\% and 9.96\% EERs) compared to the TUSUR KIBEVS results (13.39\% and 11.42\% EERs).

\section{Conclusions}\label{Conclusions}
This paper has described the experimental framework and results of the ICDAR 2021 Competition on On-Line Signature Verification (SVC 2021). The goal of SVC 2021 is to evaluate the limits of on-line signature verification systems on popular scenarios (office/mobile) and writing inputs (stylus/finger) through large-scale public databases. The following tasks are considered in the competition: \textit{i)} Task 1, analysis of office scenarios using the stylus as input; \textit{ii)} Task 2, analysis of mobile scenarios using the finger as input; and \textit{iii)} Task 3, analysis of both office and mobile scenarios simultaneously. In addition, both random and skilled forgeries are simultaneously considered in each task in order to simulate realistic scenarios.

The results achieved in the final evaluation stage of SVC 2021 have proved the high potential of deep learning methods compared to traditional approaches such as Dynamic Time Warping (DTW). In particular, the winner of SVC 2021 has been the DLVC-Lab team that proposed an end-to-end trainable deep soft-DTW  (DSDTW). The results achieved in terms of Equal Error Rates (EER) are 3.33\% (Task 1), 7.41\% (Task 2), and 6.04\% (Task 3). These results prove the challenging conditions of SVC 2021 compared to previous international competitions~\cite{yeung2004svc2004,conf/icdar/BlankersHFV09,Houmani2012993,malik2013icdar,malik2015icdar2015}, specially for the mobile scenario (Task 2).

SVC 2021 will be established as an on-going competition\footnote{\url{https://sites.google.com/view/SVC2021}}, where researchers can play fair by benchmarking easily their systems against the state of the art in an open common platform using large-scale public databases such as DeepSignDB\footnote{\url{https://github.com/BiDAlab/DeepSignDB}} and SVC2021\_EvalDB\footnote{\url{https://github.com/BiDAlab/SVC2021\_EvalDB}}, and standard experimental protocols. 

\section*{Acknowledgments}
This work has been supported by projects: PRIMA (H2020-MSCA-ITN-2019-860315), TRESPASS-ETN (H2020-MSCA-ITN-2019-860813), BIBECA (RTI2018-101248-B-I00 MINECO/FEDER), Orange Labs, and by UAM-Cecabank. 



%
%
%
 \bibliographystyle{splncs04}
 \bibliography{egbib2}
%
%
%
%
%

\end{document}